\documentclass{article} % For LaTeX2e
\usepackage{iclr2023_conference,times}
\usepackage{graphicx}
\usepackage{amsmath}
\usepackage{array}
\usepackage{multirow}
\usepackage{booktabs}
\usepackage{rotating}
\usepackage{caption}
\usepackage{subfigure}
\usepackage{adjustbox}
\usepackage{longtable}
\usepackage{hyperref}
\usepackage{url}
\usepackage{natbib}

\usepackage{amsmath,amsfonts,bm}

\def\eqref#1{equation~\ref{#1}}
\def\1{\bm{1}}

\DeclareMathAlphabet{\mathsfit}{\encodingdefault}{\sfdefault}{m}{sl}
\SetMathAlphabet{\mathsfit}{bold}{\encodingdefault}{\sfdefault}{bx}{n}

\renewcommand{\cite}{\citep}

\title{\includegraphics[width=0.03\textwidth]{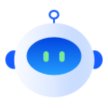} OmChat: A Recipe to Train Multimodal Language Models with Strong Long Context and Video Understanding}

\author{Tiancheng Zhao, Qianqian Zhang, Kyusong Lee, Peng Liu, Lu Zhang,  \\ \textbf{Chunxin Fang}, \textbf{Jiajia Liao}, \textbf{Kelei Jiang}, \textbf{Yibo Ma} \textbf{and Ruochen Xu} \\
Linker Technology Research\\
Binjiang Institute of Zhejiang University\\
\texttt{\{tianchez\}@hzlh.com} 
}

\iclrfinalcopy % Uncomment for camera-ready version, but NOT for submission.
\begin{document}

\maketitle

\begin{abstract}
We introduce OmChat, a model designed to excel in handling long contexts and video understanding tasks. OmChat's new architecture standardizes how different visual inputs are processed, making it more efficient and adaptable. It uses a dynamic vision encoding process to effectively handle images of various resolutions, capturing fine details across a range of image qualities. OmChat utilizes an active progressive multimodal pretraining strategy, which gradually increases the model’s capacity for long contexts and enhances its overall abilities. By selecting high-quality data during training, OmChat learns from the most relevant and informative data points.
With support for a context length of up to 512K, OmChat demonstrates promising performance in tasks involving multiple images and videos, outperforming most open-source models in these benchmarks. Additionally, OmChat proposes a prompting strategy for unifying complex multimodal inputs including single image text, multi-image text and videos, and achieving competitive performance on single-image benchmarks.
To further evaluate the model's capabilities, we proposed a benchmark dataset named Temporal Visual Needle in a Haystack. This dataset assesses OmChat's ability to comprehend temporal visual details within long videos.
Our analysis highlights several key factors contributing to OmChat's success: support for any-aspect high image resolution, the active progressive pretraining strategy, and high-quality supervised fine-tuning datasets. This report provides a detailed overview of OmChat's capabilities and the strategies that enhance its performance in visual understanding.

\end{abstract}
\section{Introduction}
In recent years, the ability to process and understand multimodal data has become increasingly critical for developing advanced AI systems~\cite{liu2024visual,liu2024llava, bai2023qwen, ai2024yi}. Models that can handle both textual and visual inputs are essential for a wide range of applications, from video analysis to complex image processing tasks. One of the key challenges in this domain is efficiently managing and leveraging long-context data, which includes sequences of images and video frames that can span significant temporal lengths.

To address these challenges, we introduce OmChat, a strong and efficient model designed to excel in handling long multimodal contexts and understanding video data. OmChat employs an active progressive multimodal pretraining strategy, which gradually scales the model’s capacity for processing long contexts and enhances its overall capabilities. By selectively utilizing high-quality data during training, OmChat is able to learn from the most relevant and informative data points, ensuring robust performance across various tasks.

OmChat supports a context length of up to 512K tokens, making it well-suited for tasks involving multiple images and videos. In benchmarks for these tasks, OmChat consistently outperforms all other open-source models, demonstrating its superior ability to manage and interpret complex visual data. Additionally, OmChat 8B model achieves competitive performance on single-image benchmarks, often surpassing larger size models.

To further evaluate OmChat's capabilities, we proposed a benchmark dataset named Temporal Visual Needle in a Haystack. This dataset is designed to assess the model's ability to comprehend and process temporal visual details within videos, challenging OmChat to locate and interpret key information embedded within long video sequences.

Our analysis identifies several key factors that contribute to OmChat's success, including the support for higher image resolutions, the active progressive pretraining strategy, and the incorporation of high-quality supervised fine-tuning (SFT) datasets. These elements collectively enhance OmChat's efficiency, adaptability, and overall performance in visual understanding tasks.

In this paper, we provide a comprehensive overview of OmChat's capabilities and the innovative strategies. We detail the model's architecture, training methodology, and performance across various benchmarks. Our findings highlight the importance of higher image resolutions, progressive multimodal pretraining, and high-quality data selection in achieving state-of-the-art performance in multimodal large language models.

The structure of this paper is as follows: Section 2 details the overall architecture and training methods of OmChat, including the vision tower and dynamic vision encoding processes. Section 3 presents the training data recipe and Section 4 shows the results of our evaluations on single-image, multi-image, and video benchmarks. Section 5 discusses the ablation and analysis.

\section{Method}
%OmChat is designed to support various context lengths to cater to different task requirements. We offer three model sizes with context capabilities of 32K, 128K, and 512K tokens. These extensive context lengths are crucial for tasks that require processing and understanding long sequences of data, such as video analysis and multi-image instructions. 

% The base model is designed for high efficiency and scalability, featuring 8 experts that dynamically activate depending on the input, optimizing computational resources. This MoE architecture allows the model to handle diverse and complex tasks more effectively by distributing the workload across multiple experts. 

\begin{figure}[ht!]
  \centering
  \includegraphics[width=13cm]{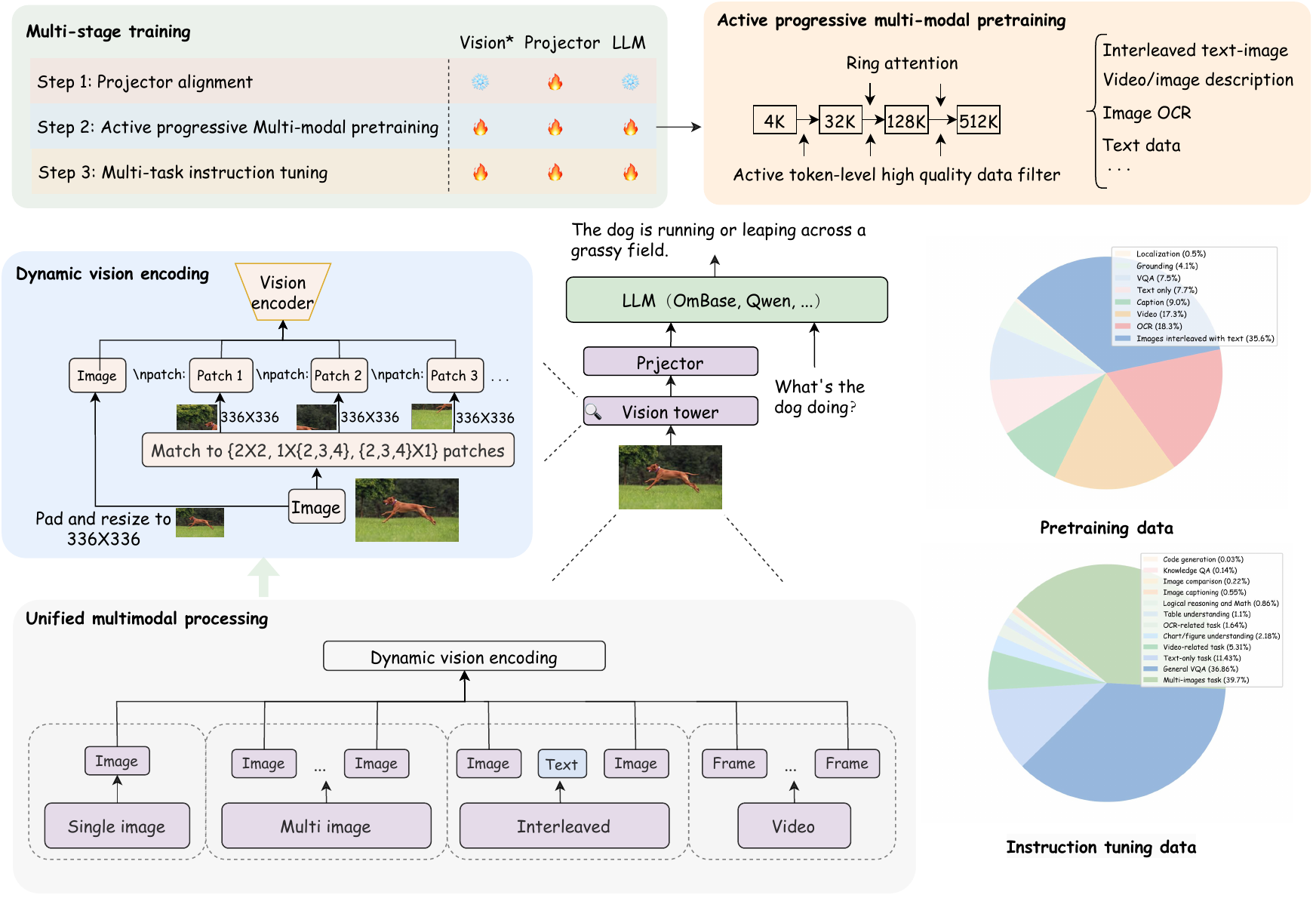}
  \caption{\textbf{OmChat model overall structure and training steps.}}
  \label{structure}
\end{figure}

The overall OmChat architecture and training method is depicted in Figure \ref{structure}. OmChat processes both visual and textual data inputs. In the context of multimodal tasks, the visual input can vary from a single image to multiple images, image-text interleaved data, and video frames. Conversely, for language tasks, the visual component may be absent. These diverse inputs are subsequently fed into a large language model for processing. Ultimately, the output generated by OmChat manifests in textual form. 

\textbf{Unified Multimodal Processing} OmChat implements a unified approach to processing various types of visual inputs. Regardless of the input format, OmChat standardizes the procedure by first decomposing the inputs into images before channeling them into the vision tower. This systematic method ensures that all input variations, whether they involve single images, multiple images, image-text combinations, or video frames, undergo a consistent transformation process. This unified process not only enhances the model's efficiency in handling different types of visual data but also underscores the model's adaptability and robustness in accommodating a wide range of input modalities. 

\textbf{Dynamic Vision Encoding:} In order to effectively address images with varying resolutions, OmChat has implemented a dynamic vision encoding process inspired by AnyRes~\cite{liu2024llavanext}. Our innovative approach ensures that the model can adeptly handle images of different resolutions without overlooking small objects that may be present in high-resolution images. By incorporating this dynamic vision encoding mechanism, OmChat enhances its capability to capture fine details and nuances across a spectrum of image resolutions, thereby improving the overall accuracy and robustness of its vision capabilities. 

\textbf{Multi-Stage Training:} OmChat's training process unfolds in three distinct steps to optimize its capabilities effectively. During the initial phase, the vision tower and the language model remain frozen. The focus is on training the projector that bridges the visual and textual modalities. By isolating this component for training, OmChat optimizes the connections between vision and text inputs, ensuring seamless integration and effective communication between the two domains. The second step involves multimodal generative training, where both the vision encoder, the language model and the projector are optimized. In this stage, the training objective revolves around minimizing the cross-entropy of the text tokens. By updating the vision encoder, the language model and the projector simultaneously, OmChat enhances its ability to generate coherent and contextually relevant responses across different modalities. This comprehensive training approach strengthens the model's multimodal understanding and generation capabilities.

\textbf{Active Progressive Multimodal Pretraining:} We implement a progressive training strategy from 4K to 512K to scale up the model's long context capacity and capabilities gradually. Additionally, the implementation of high-quality selection during training is a crucial step in ensuring that the model learns from the most informative and relevant data points. 

We further detail the key components essential for enhancing OmChat: 1) support for high resolutions with dynamic vision encoding, 2) a progressive training strategy for long contexts, and 3) the selection of high-quality instruction tuning data.

\subsection{Dynamic Vision Encoding}

The vision encoder is a crucial component of multimodal systems. Previous research has demonstrated that supporting various resolutions can lead to significant improvements in multimodal training~\cite{liu2024llavanext,dong2024internlm,she2024mammothmoda}. Our findings also show that dynamic vision encoders greatly enhance performance. Additionally, we employ specific data formats and delimiters to differentiate between image patches and various types of visual inputs. For example, a single image is processed as an individual entity, while videos are treated as sequences of frames. Delimiters mark the beginning and end of each frame in a video sequence, enabling the model to effectively understand and process the temporal aspects of video data. 

We adopted the AnyRes technique, which enables our vision tower to support images and videos of any resolution. AnyRes dynamically adjusts the processing pipeline to handle varying resolutions, ensuring that the model can process high-resolution inputs efficiently and accurately. Our dynamic image encoding method is based on AnyRes. Additionally, we incorporate delimiters to help the model differentiate between patches, images, and video frames, thereby enhancing its ability to understand dynamic visual inputs.

% The vision tower is employed to process images using a grid of specific pinpoints, defined as \texttt{(768, 768), (384, 768), (768, 384), (384, 1152), (1152, 384), (768, 1152), (1152, 768)}.
% To efficiently encode the images, we first determine the most suitable image size from the predefined grid pinpoints based on the input image size. Once the optimal size is selected, the image is converted into n patches along with one whole image. The whole image is resized to a 
% 384×384 dimension and subsequently encoded using the vision tower.
% The resulting encoding from the whole image is combined with the encoded patches. 
% These patches are individually encoded and concatenated with a determiner ``patch:" along with a newline character (\textbackslash n) inserted between each patch embedding. We discovered that the sequence and format of prompts significantly impact performance on certain tasks. Based on the preliminary study, We have established the prompt format for our model for inputs containing a single image as follows:

% \begin{verbatim}
% <image>\n<patch>\n...\n<patch>\n{text}
% \end{verbatim}

% In this format, \texttt{<image>} denotes the embedding of the entire image, \texttt{<patch>} represents the embedding of individual image patches, and \texttt{\{text\}} is the textual prompt. This structured prompt format ensures a coherent integration of visual and textual information, facilitating effective processing and interpretation by the model.
% % We utilize the \texttt{siglip-so400m-patch14-384} model. 

\subsection{Progressive Training Strategy to Long Context}

To enhance the model's ability to process longer contexts effectively, a progressive training strategy is implemented \citep{liu2024worldmodelmillionlengthvideo}. Initially, the language base model is expanded to 512K by leveraging a text pretraining dataset comprising diverse source data. This expansion builds upon our original language model, i.e., OmBase. The training strategy follows a sequential context length of 4k, 32K, 128K, and finally 512K. Additionally, we also train OmChat using Qwen2-7B \citep{qwen}, which initially enables a context length of 32k. In this case, we extend the context length to 128K and then 512K. By extending the context length successively, OmChat retains its proficiency in processing short contexts while developing the capacity to handle longer contexts at a relatively lower cost. Notably, the positional encoding parameters are adjusted by scaling up the RoPE  
$\theta$ to 50M. 
% In the context of constructing training data with extended contexts, a method is employed wherein training samples shorter than the designated context length for each phrase were concatenated. This approach ensures that the model could be effectively trained on the intended context. 
In situations where there is minimal or no data available for extensive context lengths, we employ a specific strategy to ensure our model is adequately trained. This involves the concatenation of shorter samples to generate samples that align with the desired context length.
Subsequently, leveraging the language model with a 512k context length, omchat is transformed into a multimodal model through a multimodal pretraining process. After the projector alignment training, the model's context length was progressively extended by incrementally increasing the context length from 4K, 32K, 128K, to 512K. Details regarding the training data are elaborated on in Subsection \ref{train_data}, where, akin to text pretraining, training samples that are shorter than the intended context are concatenated to form a single sample. The RoPE $\theta$ is maintained at 50M.

During the training phase for contexts exceeding 128K in length, RingAttention \citep{liu2023ring} is implemented for computing QKV attention. This specialized method is tailored for calculating attention over long contexts, addressing memory constraints associate with the quadratic complexity of attention weight computations. The fundamental concept involves segmenting QKV along the \texttt{seq\_len} dimension into blocks of \texttt{n*block\_size}, facilitating the iterative derivation of complete attention by calculating attention for each block.

\subsection{High-Quality multimodal Pretrain Data Selection }
Enhancing model performance through multimodal pretraining heavily relies on the quality of the training dataset.
In this pursuit, we leverage an innovative technology known as Rho-1 \cite{lin2024rho} for the purpose of high-quality data selection.
% Rho-1 scrutinizes the dataset to pinpoint the top 10% of data samples that exhibit the highest representativeness and value for training.
Expanding upon the original Rho-1 methodology, initially tested in the context of language-only pretraining, we extend its application to the realm of multimodal fine-tuning. In alignment with our perspective that ``Not all tokens in a corpus hold equal importance for multimodal training," tokens within multimodal data can be categorized into three distinct types \cite{xiao2024seeing}:
\begin{itemize}
    \item Type 1: Text highly related to images: entities (e.g., people, animals, objects), quantities, colors, text, etc. These tokens directly correspond to image information and are crucial for multimodal alignment.
    \item Type 2: Text with low relevance to images: transitional words or content that can be inferred from the preceding text. These tokens primarily serve to train the pure text capabilities of MLLM.
    \item Type 3: Text conflicting with image content: These tokens are inconsistent with image information, potentially providing misleading information and negatively impacting the multimodal alignment process.
\end{itemize}

We propose a Selective Visual Language Modeling (SVLM), to prioritize type 1 text and disregard type 3 text. SVLM starts with training a reference model on high-quality multimodal instruction tuning data. It then computes the reference loss for all text tokens in multimodal pretraining based on the logarithmic probability derived from the reference model. To distinguish the three types of tokens, we generalize the excess loss in Rho-1, which is the difference between the pretrain loss and the reference loss, to a multimodal setting. By selectively retaining tokens with high excess loss, the focus can be efficiently directed towards type 1 tokens while filtering out type 3 tokens. 
We compute the reference loss offline for the multimodal corpus and incorporate it into the batched data for real-time computation of excess loss during training. During step-2 multimodal pretraining, we prioritize tokens based on their excess loss values within a batch, focusing solely on the top percentile of tokens for loss computation. The additional steps of loading the reference loss and conducting ranking have minimal impact on the training process. 
The efficacy of SVLM is validated across six benchmarks following the second phase of multimodal pretraining. Specifically, we evaluated the performance on CMMLU \cite{li2023cmmlu}, CEval \cite{huang2024c}, GSM8K \cite{cobbe2021gsm8k}, MATH \cite{hendrycks2measuring}, HumanEval \cite{chen2021codex}, and BBH \cite{suzgun2022challenging}. The average score exhibited an improvement from 32.7 to 38 after the implementation of SVLM.

\subsection{High-quality Instruction Tuning Data Selection}
\begin{figure}[htb]
  \centering
  \includegraphics[width=10cm, height=7cm]{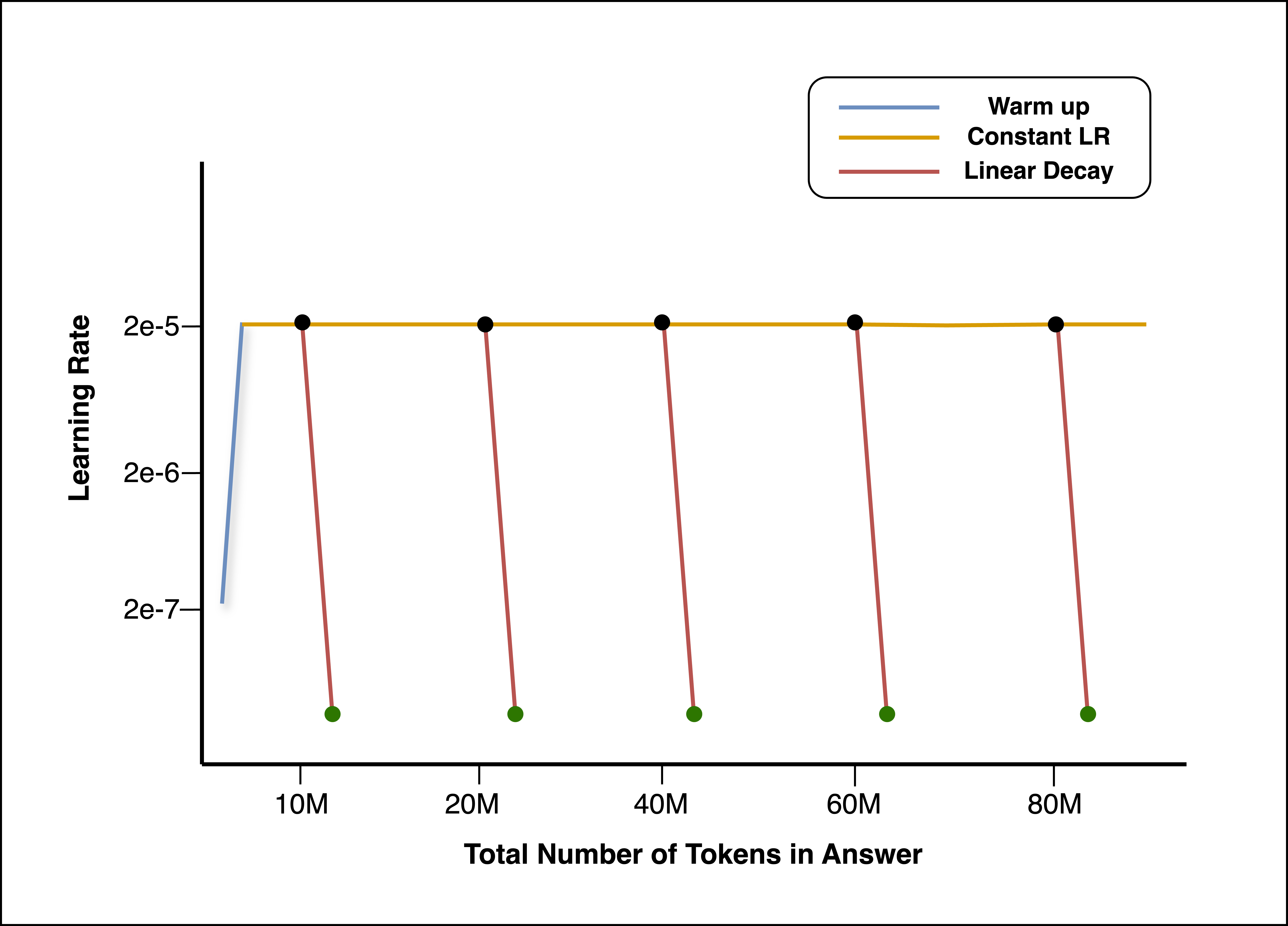}
  \caption{\textbf{Illustration of the continuous training strategy.} The black dot represents the step at which a checkpoint is saved using a constant learning rate for the current data portion. Based on each of these checkpoints and the corresponding data portions, we perform additional training for an extra 10\% of the steps on the same data using linear decay scheduler, as indicated by the red lines. The green dot represents the checkpoint used to evaluate the performance of this data portion.}
  \label{continuous_training}
\end{figure}
To enhance the multimodal capabilities of our model, we meticulously curate over 60 datasets from a diverse array of vision-language tasks for the fine-tuning stage. These datasets encompass a broad spectrum of key areas, including general visual question answering, OCR-related tasks, chart and diagram understanding, image quality assessment, image and video captioning, document-related queries, mathematical and computational tasks, multi-image analysis, and video comprehension, among others. From another perspective, these datasets can be classified into four categories: text-only tasks, single-image tasks, multi-image tasks, and video-related tasks. Unlike other models, we have specifically included tasks that utilize multi-image or video data, such as image comparison and video question answering. This approach helps the model learn to comprehend multi-image scenarios and the temporal relationships among video frames, thereby enhancing its performance in multi-image and video understanding.

One challenge of using these datasets for fine-tuning is the disparity in their sizes. For instance, OpenHermes-2.5 contains approximately 1 million samples, whereas InfoVQA has only 2,000 samples. Directly using all datasets may lead to an imbalance in task variety, making it crucial to find an optimal approach for data combination. Inspired by the findings in \cite{hu2024minicpm}, we employ a continuous training strategy to select the best instruction tuning data combination. As shown in Figure \ref{continuous_training}, we first manually design the proportion of each dataset in percentage based on the importance of the tasks. Then, based on the total number of tokens for the answers, we sample a large combination from the original datasets according to the designed percentages. For datasets with insufficient data, we perform repeated sampling to meet the required amount. Afterward, we divide this a large combination into multiple parts, ensuring that each part maintains the designed percentages. Next, we apply a continuous training strategy. The model is initially trained on the first portion of the large data combination using a constant learning rate of 2e-5 with warm up. Subsequently, the second portion of the large data combination is trained starting from the last checkpoint of the first portion using the same constant learning rate, and this process continues for the remaining portions. At the same time, we take the last checkpoint from each portion and perform additional training on the same data used in constant phase, followed by a linear decay scheduler for an extra 10\% of the total steps. In this way, we are able to evaluate the performance of instruction tuning datasets at different scales without repeatedly training the previous portions. This allows us to select the best combination of portions from all the instruction tuning datasets. Detailed results are presented in Subsection \ref{data selection ablation}.

\section{Training Data Recipe}
% \subsection{Training Data} 
The utilisation of diverse and large-scale datasets represents a pivotal component in the training of multimodal models. Our datasets are categorized into two main parts: pretraining data and instruction tuning data. The pretraining data includes both vision-text and text-only data from various sources, aimed at enhancing the model's foundational cross-modal understanding and knowledge. Besides, the instruction tuning data is smaller in scale and is primarily used to train the model for specific downstream tasks, further refining and optimizing its performance in real-world applications. This two-stage data construction approach ensure that the model not only possesses a broad range of universal capabilities but also excels in specific tasks.

\subsection{Pretraining data} \label{train_data}
The pretraining data used in our study encompasses a variety of publicly available sources along with some proprietary data. The pie chart to the right of Figure \ref{structure} visually represents these data as a percentage of the whole.

Specifically, we divide the whole pretraining dataset into the following categories:

\begin{itemize}
    \item \textbf{Image interleaved with text}: This type of data enables the model to improve its contextual learning for multimodal inputs. We use MMC4-Core \citep{zhu2024multimodal} and in-house interleaved data (e.g. news data), which help the model understand and process contexts where images alternate text.
    
    \item \textbf{Optical Character Recognition (OCR)}: These data enhance the model's OCR capabilities at both the document and image levels. Publicly available datasets including CTW \citep{yuan2019large}, LSVT \citep{sun2019chinese}, ReCTS \citep{yao2012detecting}, ICDAR2019-ArT \citep{chng2019icdar2019}, and pdfa-eng-wds \citep{pixparse2024} are used in the pretraining stage.
    
    \item \textbf{Video}: This category of data empowers the model to comprehend the sequence of multiple images and frames, thereby facilitating the processing of continuous actions and events in videos. We utilize public datasets such as InternVideo \citep{wang2022internvideo} and ActivityNet \citep{caba2015activitynet}, as well as various in-house video data from sources like movies and news. The appropriate frame rate for frame extraction is selected based on the duration of each video, with frame rates ranging from 0.5 FPS to 30 FPS.
    
    \item \textbf{Caption}: The data used consist of several public datasets and a few in-house datasets. The major public datasets include CC12M \citep{changpinyo2021conceptual}, ShareGPT4V \citep{chen2023sharegpt4v}, VizWiz-Captions \citep{gurari2020captioning}, SBU \citep{ordonez2011im2text}, Flickr30K \citep{young2014image}, Microsoft COCO Captions \citep{chen2015microsoft}, Taisu \citep{liu2022taisu}, ALLaVA-4V \citep{chen2024allava}, Laion-400M \citep{schuhmann2021laion}, CC3M \citep{sharma2018conceptual}, TextCaps \citep{sidorov2020textcaps}. These datasets assist the model in creating associations between images and descriptive text.
    
    \item \textbf{Text-only}: These datasets, which include RedPajama \citep{together2023redpajama}, Alpaca \citep{taori2023stanford}, Belle \citep{BELLE},  DeepCtrl-sft-data \citep{DeepCtrl-sft-data}, Wudao \citep{Wudao}, Telechat \citep{wang2024telechat}, Firefly \citep{Firefly}, pCLUE \citep{pCLUE}, Guanaco \citep{Guanaco}, are primarily utilized to enhance the language capabilities of our model.
    
    \item \textbf{Visual Question Answering (VQA)}: This category of datasets includes VQAv2 \citep{goyal2017making}, M3IT \citep{li2023m}, VizWiz-VQA \citep{gurari2018vizwiz}, OCR-VQA \citep{mishra2019ocr}, Visual Spatial Reasoning (VSR) \citep{liu2023visual}, OK-VQA \citep{marino2019ok}, Visual Dialog (VisDial) \citep{das2017visual}, A-OKVQA \citep{schwenk2022okvqa}, TextVQA \citep{singh2019towards}, GQA \citep{hudson2019gqa}, along with several in-house VQA datasets. These datasets facilitate the model handle tasks that require joint visual and textual reasoning.
    
    \item \textbf{Grounding and Localization}: To guide our model to better interpret and interact with the visual world, we incorporate grounding and localization datasets during the pretraining stage. The grounding data are sourced from high-quality in-house data and GoldG~\cite{li2022grounded}, and localization data primarily derived from RefCOCO, RefCOCO+, and RefCOCOg \citep{yu2016modeling}. These datasets enable the model to accurately localize and identify specific objects and regions within images.
\end{itemize}

To ensure the quality of training during the pretraining stage, we implement a series of data cleaning procedures on the pretraining datasets. Firstly, sensitive information is meticulously screened and removed to safeguard privacy and comply with relevant regulations. Secondly, we apply data deduplication techniques to eliminate redundant entries, thereby enhancing the diversity and effectiveness of the dataset. Additionally, we conduct thorough damage detection and cleaning of image data, removing corrupted images to ensure the model receives high-quality inputs during pretraining. These comprehensive cleaning operations not only improve the purity and reliability of the dataset but also establish a robust foundation for subsequent model training.

% \begin{figure}[htb]
%   \centering
%   \includegraphics[width=8cm]{pre_train.jpg}
%   \caption{Distribution of data category used in the multimodal pretraining stage}
%   \label{fig: pre_train}
% \end{figure}

\subsection{Instruction tuning data}

The instruction tuning data employed in our study encompass a wide range of vision-text and text-only datasets, addressing a variety of challenging downstream tasks.  Additionally, the pie chart on the right side of the Figure \ref{structure} visually represents the proportion of each of task data within the whole dataset.

To be more precise, we divide the entire dataset into the following categories based on downstream tasks:

\begin{itemize}
    \item \textbf{Multi-images task}: This type of tasks focuses on visual question answering involving multiple images. For this purpose, we use the Mantis-Instruct \citep{jiang2024mantis}, which covers a diverse array of multi-image skills such as co-reference, reasoning, comparing, temporal understanding.
    
    \item \textbf{General VQA}: The primary distinction from the multi-images task is that general VQA typically involves questions and answers pertaining to individual images. For this task, we use datasets such as COCO-QA \citep{ren2015exploring}, Visual7W \citep{zhu2016visual7w}, VQAv2 \citep{goyal2017making}, TallyQA \citep{acharya2019tallyqa}, HatefulMemes \citep{kiela2020hateful}, VQA-RAD \citep{lau2018dataset}, LLaVA-Instruct-150K (llava\_v1\_5\_mix665k) \citep{liu2024improved}, M3IT \citep{li2023m}, and ALLaVA-Instruct-4V \citep{chen2024allava}.
    
    \item \textbf{Text-only task}: This task involves handling not only general textual dialogues but also mathematical problems and arithmetic calculations. For this purpose, we primarily rely on publicly available datasets including OpenHermes-2.5 \citep{OpenHermes_2.5}, Dolly \citep{conover2023free}, MetaMathQA \citep{yu2023metamath}, MathInstruct \citep{yue2023mammoth}, CamelAIMath \citep{li2023camel}, AtlasMathSets \citep{AtlasMathSets}, Goat \citep{liu2023goat}, CoT \citep{alpaca-cot}, pCLUE \citep{pCLUE}, Firefly \citep{Firefly}, COIG \citep{zhang2023chinese}, and Alpaca \citep{taori2023stanford}.
    
    \item \textbf{Video-related task}: This task primarily designed to enable the model to understand videos. The specific datasets used include Video Instruction \citep{Maaz2023VideoChatGPT}, InternVideo \citep{wang2022internvideo}, and Video-ChatGPT-100K \citep{Maaz2023VideoChatGPT}. Note that we select the appropriate frame rate (ranging from 0.5FPS to 30FPS) based on the duration of each video for frame extraction, so that videos of different length have similar number of frames. 
    % While there is overlap between Video Instruction and Video-ChatGPT-100K, the frame extraction frequencies for these videos are different.

    \item \textbf{Chart/figure understanding}: The data utilized for this type of task are all derived from publicly available datasets, including Chart2Text \citep{obeid2020chart}, DVQA \citep{kafle2018dvqa}, ChartQA \citep{masry2022chartqa}, FigureQA \citep{kahou2017figureqa}, MapQA \citep{chang2022mapqa}, and MMC-Instruction \citep{liu2023mmc}.

    \item \textbf{OCR-related task}: The scope of this task extends beyond mere OCR to encompass document understanding and text transcription. The datasets employed in our study include RenderedText \citep{RenderedText}, DocVQA \citep{mathew2021docvqa}, TextVQA \citep{singh2019towards}, ST-VQA \citep{biten2019scene}, VisualMRC \citep{tanaka2021visualmrc}, IAM \citep{marti2002iam}, InfoVQA \citep{mathew2022infographicvqa}, Diagram image-to-text \citep{image-to-text}, RCTW-17 \citep{shi2017icdar2017}, and ReCTS \citep{yao2012detecting}.

    \item \textbf{Table understanding}: The datasets utilized for this type of task comprise TabMWP \citep{lu2022dynamic}, TAT-QA \citep{zhu2021tat}, HiTab \citep{cheng2021hitab}, MultiHiertt \citep{zhao2022multihiertt}, FinQA \citep{chen2021finqa}, WikiSQL \citep{zhong2017seq2sql}, SQA \citep{iyyer2017search}, and WTQ \citep{pasupat2015compositional}.

    \item \textbf{Logical reasoning and Math}: This part of datasets covers GeomVerse \citep{kazemi2023geomverse}, CLEVR-Math \citep{lindstrom2022clevr}, CLEVR \citep{johnson2017clevr}, IconQA \citep{lu2021iconqa}, RAVEN \citep{zhang2019raven}, and Inter-GPs \citep{lu2021inter}.

    \item \textbf{Image captioning}: As one of the prevalent downstream tasks for multimodal models, we primarily utilize the LNarratives \citep{pont2020connecting}, Screen2Words \citep{wang2021screen2words}, ShareGPT4V \citep{chen2023sharegpt4v}, and some in-house caption data.

    \item \textbf{Image comparison}: This task involves employing models to compare two images and analyze the differences between them. Public datasets such as NLVR2 \citep{suhr2018corpus}, GSD \citep{li2023mimic}, and Spot the diff \citep{jhamtani2018learning} are utilized for this purpose.

    \item \textbf{Knowledge question answering (Knowledge QA)}: This task involves using models to answer intellectual questions originated from textbooks or academic texts. For this purpose, We utilize AI2D \citep{kembhavi2016diagram}, TQA \citep{kembhavi2017you}, and ScienceQA \citep{lu2022learn}.

    \item \textbf{Code generation}: As the name implies, this task focuses on the utilization of models to generate code. We select WebSight \citep{laurenccon2024unlocking} and DaTikz \citep{belouadi2023automatikz} as the primary datasets for this task.
\end{itemize}

To avoid data contamination, rigorous processing and validation are conducted to guarantee that the data in the evaluation benchmarks is entirely excluded from the pretraining and instruction tuning stages. This strict separation ensures no overlap between the training and test datasets, upholding the fairness and credibility of the comparison results.

% \begin{figure}[htb]
%   \centering
%   \includegraphics[width=8cm]{sft.jpg}
%   \caption{Distribution of data category used in the multimodal supervised instruction tuning stage}
%   \label{fig: pre_train}
% \end{figure}

% \subsection{The Construction of Video Needle-in-the-Haystack}

\section{Experiments}
\subsection{Benchmarks for Evaluation}

We evaluate our models using a variety of publicly available multimodal benchmarks to assess the performance across different tasks.

1. General Single-image Datasets:
\begin{itemize}
    \item \textbf{MMBench V1.1} \citep{MMBench}\textbf{ (MMBench-CN v1.1 and MMBench-EN v1.1)} is a comprehensive benchmark includes over 3,000 multiple-choice questions covering 20 distinct ability dimensions, including object localization, social reasoning, and image emotion recognition. It evaluates cognitive abilities and linguistic competencies in both Chinese and English.
    \item \textbf{MMStar} \citep{chen2024we} evaluates MLLM's abilities in Coarse Perception (CP), Fine-grained Perception (FP), Logical Reasoning (LR), Instance Reasoning (IR), Science \& Technology (ST), and Mathematics (MA). It includes 1,500 challenging samples selected by humans, evaluating the model's understanding and reasoning with visual content across different complexity levels.
    \item \textbf{MMMU} \citep{yue2023mmmu} probes multi-disciplinary competencies with 11.5K curated questions from college resources. Covers Art \& Design, Business, Science, Health \& Medicine, Humanities \& Social Science, and Tech \& Engineering. Includes diverse question types for comprehensive evaluation.
    \item \textbf{HallusionBench} \citep{guan2023hallusionbench} assesses the language hallucination and visual illusion of MLLMs with 455 visual-question control pairs.  It includes 346 unique figures and a total of 1129 questions spanning diverse topics and formats.
    \item \textbf{AI2D} \citep{kembhavi2016diagram} is a dataset for understanding science diagrams, consisting of over 5,000 diagrams representing grade school science topics. Each diagram is annotated with constituent segmentations, their relationships to each other, and their relationships to the diagram canvas. 
     \item \textbf{MMVet}\citep{yu2024mm} evaluates a wide range of abilities, including recognition, OCR, knowledge retention, language generation, spatial awareness, and mathematical proficiency. It features 16 tasks for quantitative evaluation and includes a curated collection of 200 images and 218 samples, each paired with their corresponding ground truths. MMVet integrates to solve various complex multimodal tasks.
\end{itemize}
2. Math Reasoning Datasets:
\begin{itemize}
    \item \textbf{MathVista} \citep{lu2024mathvista} is a comprehensive benchmark that evaluates a model's understanding of mathematical and visual tasks by combining 28 multimodal datasets, comprising 9 MathQA datasets and 19 VQA datasets. It introduces three new datasets (IQTest, FunctionQA, PaperQA) to assess logical reasoning through puzzle test figures, algebraic reasoning with functional plots, and scientific reasoning using academic paper figures.
\end{itemize}
    
3. OCR-related Image Understanding Datasets:
\begin{itemize}
    \item \textbf{OCRBench} \citep{liu2024hidden} is a comprehensive OCR evaluation benchmark featuring 29 datasets and five key components: text recognition, Scene Text-Centric VQA, Document-Oriented VQA, KIE, and HMER. It includes 1000 question-answer pairs, making it a thorough assessment of a model's OCR capabilities.
\end{itemize}

4. General Multi-image Datasets:
\begin{itemize}
    \item \textbf{Mantis-Eval} \citep{jiang2024mantis} is a challenging dataset featuring 217 multi-image reasoning examples covering various skills such as size perceptions and weight comparisons. The dataset includes both multiple-choice and short-answer questions. 
    \item \textbf{Q-Bench}\citep{wu2024qbench} assesses the capability of MLLMS in evaluating and comparing visual quality. It focuses on testing MLLMs' low-level visual skills, particularly their ability to assess image quality. The evaluation is conducted on the Qbench2-A2-pair development set, which presents 1000 multiple-choice questions based on various image contents.
    \item \textbf{MileBench} \citep{song2024milebench} rigorously evaluates MLLMs across diverse challenges. It comprises two evaluation sets: a diagnostic evaluation emphasizing long-context recall tasks like needle-in-a-haystack and image retrieval, and a realistic evaluation simulating real-world conditions with temporal and semantic multi-image tasks. 
    %Additionally, it allows for setting the \textit{max\_context\_len} parameter based on the model's %input capacity to vary the number of images retrieved in the needle-in-a-haystack task.
\end{itemize}

5. General Video Understanding Datasets:
\begin{itemize}
    \item \textbf{MVBench} \citep{li2023mvbench} is a varied benchmark for multimodal video understanding, encompassing 20 intricate video tasks that demand analysis of image sequences for precise solutions, rather than relying solely on a single image. 
    %Additionally, it allows for the extraction of varying numbers of frames as inputs to the evaluation code, depending on experimental requirements. 
\end{itemize}

\subsection{Single Image Results}
As shown in Table \ref{tab:mmmu_test_performance}, we evaluated our model on the OpenCompass multimodal benchmarks~\cite{2023opencompass}, including MMBench, MMStar, MMMU, MathVista, etc. We also compared the evaluation scores of OmChat to other MLLM models on OpenCompass leaderboard. Our OmChat 8B model demonstrates promising results in single image inference, outperforming much larger models such as LLaVA-Next-Yi-34B~\cite{liu2024llavanext}, 360VL-70B~\cite{360vl}, CogVLM-17B-Chat~\cite{wang2023cogvlm},XVERSE-V~\cite{xverse},  IDEFICS2~\cite{laurenccon2024matters} and Yi-VL 34B~\cite{ai2024yi}. 

Other models such as MiniCPM-V2.5~\cite{hu2024minicpm} and InternLM-XComposer2-VL-4kHD (internLM-XC-HD)~\cite{dong2024internlm} also show strong performance in specific benchmarks but do not maintain consistently high scores across all tasks. Notably, OmChat achieves the highest score on MMBench (78.8), outperforming other models across several benchmarks. 

% \begin{table}[ht]
%     \centering
%     \caption{Single Image Performance}
%     \resizebox{\textwidth}{!}{% Resize table to fit within the page width
%     \begin{tabular}{lccccccccc}
%         \hline
%         \textbf{Model} & \textbf{Avg.} & \textbf{MMBench} & \textbf{MMStar} & \textbf{MMMU} & \textbf{MathVista} & \textbf{Hallusion} & \textbf{AI2D} & \textbf{OCRBench} & \textbf{MMVet} \\
%         \hline
%         GLM-4v-9B & 59.1	  & 67.9	 & 54.8  & 46.9	  & 51.1	  & 45 & 71.2	 & 776 & 58  \\
%         \hline  
%         MiniCPM-Llama3-V2.5 &58.8 & 72 & 51.8	 & 45.8  & 54.3 & 42.4 & 78.4 &725& 52.8  \\
%         \hline
%         InternLM2-7B	 & 58.8 & 76.5	 & 55.3 & 39.7 &59.4  &42.5  &81 &675 &48.2  \\
%         \hline
%         IDEFICS2-8B  & 53  & 68.9  & 49.5  & 45.2  & 52.2  & 39.1  &72.3 & 626 & 34  \\
%         \hline
%         LLaVA-Next-Mistral-7B  & 45.8  & 63.1  & 38.4  & 37	  & 34.6	  & 29.1	  &69 & 531 & 42.2  \\
%         \hline
%         Qwen-VL-Chat  & 45.2	  & 59.1	  & 34.5	  & 37	  & 34.9	  & 36.8	  & 63	 & 488 & 47.3  \\
%         \hline
%         OmChat-8B(Ours)  &   &   &   &   &   &   & &  &   \\
%         \hline
%     \end{tabular}
%     }
%     \label{tab:mmmu_test_performance}
% \end{table}

\begin{table}[ht]
    \centering
    \caption{\textbf{Single image performance.}}
    \resizebox{\textwidth}{!}{% Resize table to fit within the page width
    \begin{tabular}{lc|c|cccccccc}
        \toprule
        \textbf{Model} & \textbf{Params} & \textbf{Avg.} & \textbf{MMBench} & \textbf{MMStar} & \textbf{MMMU} & \textbf{MathVista} & \textbf{Hallusion} & \textbf{AI2D} & \textbf{OCRBench} & \textbf{MMVet} \\
        \midrule
        %GLM-4v & 9B & 59.1 & 67.9 & 54.8 & 46.9 & 51.1 & 45 & 71.2 & 776 & 58 \\
        MiniCPM-V2.5 & 8B & 58.8 & 72.0 & 51.8 & 45.8 & 54.3 & 42.4 & 78.4 & \textbf{725} & \textbf{52.8} \\
        InternLM-XC-HD & 7B & 58.8 & 76.5 & \textbf{55.3} & 39.7 & \textbf{59.4} & \textbf{42.5} & \textbf{81.0} & 675 & 48.2 \\
        %WeMM & 7B & 58.3 & 75.7 & 57 & 45.3 & 54.9 & 47.5 & 77.9 & 628 & 45 \\
        LLaVA-Next-Yi-34B & 34B & 55.0 & 77.8 & 51.6 & 48.8 & 40.4 & 34.8 & 78.9 & 574 & 50.7 \\
        IDEFICS2 & 8B & 53.0 & 68.9 & 49.5 & 45.2 & 52.2 & 39.1 & 72.3 & 626 & 34.0 \\
        XVERSE-V & 13B & 49.4 & 66.3 &49.3& 44.1& 45.3 & 33.3 & 70.6 & 489 & 37.8 \\
        360VL & 70B & 48.2 & 75.0 & 48.1 & \textbf{53.4} & 38.0 & 34.8 & 71.9 & 397 & 24.7 \\
        CogVLM-17B-Chat & 17B & 47.9 & 58.8 & 39.9 & 37.3 & 35.0 & 35.4 & 63.3 & 590 & 54.5 \\
        LLaVA-Next-Vicuna & 13B & 47.6 & 66.5 & 40.4 & 37.3 & 34.1 & 31.8 & 72.2 & 537 & 44.9 \\
        LLaVA-Next & 7B & 45.8 & 63.1 & 38.4 & 37.0 & 34.6 & 29.1 & 69.0 & 531 & 42.2 \\
        Qwen-VL-Chat &  9B & 45.2 & 59.1 & 34.5 & 37.0 & 34.9 & 36.8 & 63.0 & 488 & 47.3 \\ 
        Yi-VL & 34B & 43.5 & 67.8 & 40.5 & 45.1 & 31.5 & 35.3 & 65.9 & 290 & 32.7 \\ \midrule
        \textbf{OmChat(Ours)} & 8B & 55.9 & \textbf{78.8} &  53.8 & 45.9 & 48.3 & 40 & 77.5 & 637 & 39.6 \\
        \bottomrule
    \end{tabular}
    }
    \label{tab:mmmu_test_performance}
\end{table}

\subsection{Long Context Results}

\subsubsection{Text Needle-in-the-Haystack}

\begin{figure}[htb]
  \centering
  \includegraphics[width=10cm]{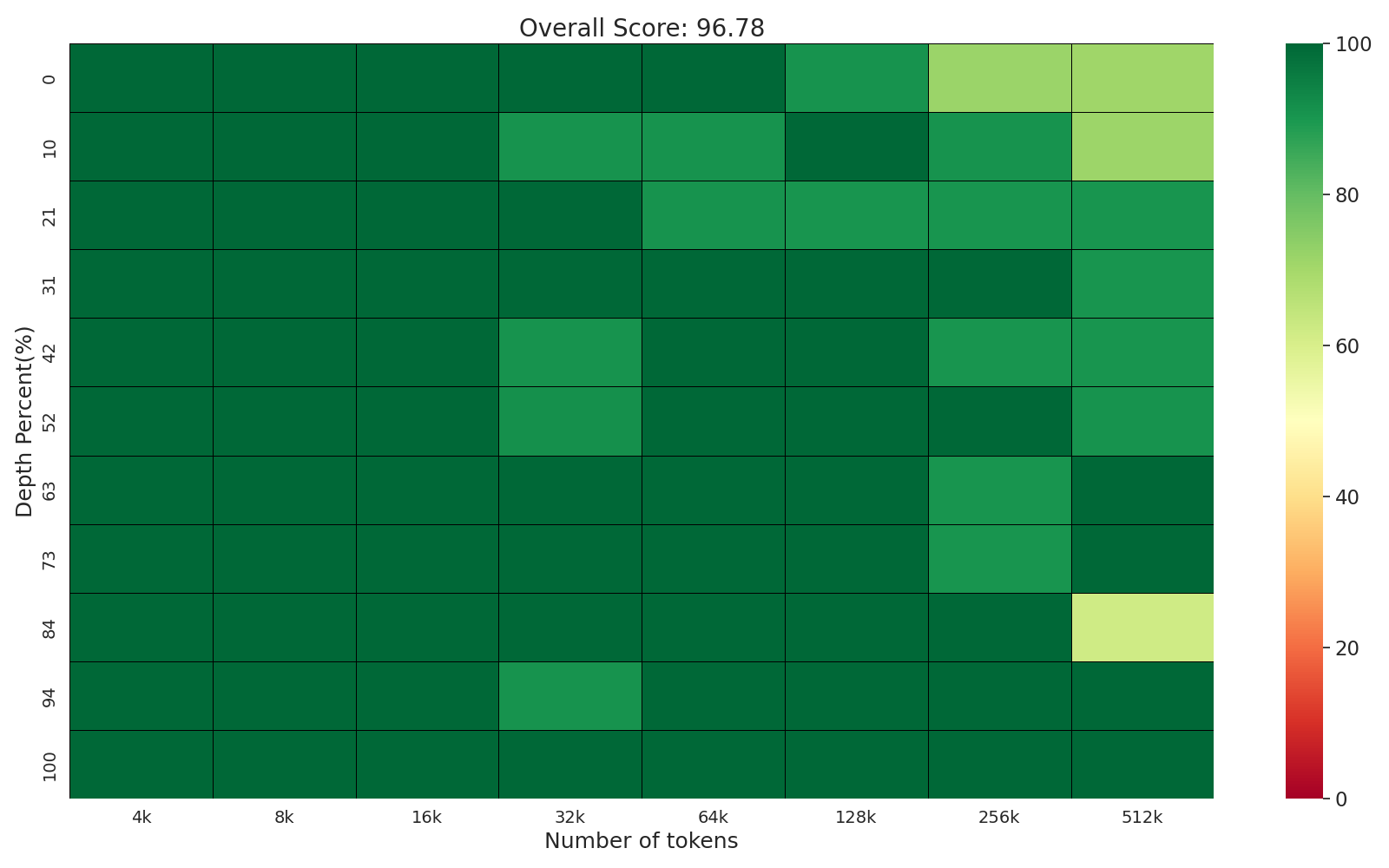}
  \caption{\textbf{Text needle retrieval performance of OmChat.}}
  \label{text_needle}
\end{figure}

As illustrated in Figure \ref{text_needle}, we evaluate our 512k OmChat model on the widely used Needle in a Haystack task \citep{NeedleInAHaystack}. Specifically, the model is evaluated with a single needle setting, where it needs to retrieve and answer a question based on a fact or statement randomly placed within a long context. OmChat demonstrates nearly perfect performance across context lengths ranging from 4k to 256k tokens, highlighting the success and effectiveness of our long context training strategy. The minor performance decrease observed at the 512k context length is likely due to fine-tuning the model with much shorter contexts. Incorporating more long context data during the fine-tuning stage should help improve this performance.

\subsubsection{Temporal Visual Needle-in-the-Haystack}

% \begin{figure}[htb]
%   \centering
%   \includegraphics[width=10cm]{video-needle_omchat.png}
%   \caption{\textbf{TV Needle performance of OmChat.}}
%   \label{text_needle}
% \end{figure}

\begin{figure}[htb]
    \centering
    \subfigure[LLaVa-1.5]{\includegraphics[width=6.8cm]{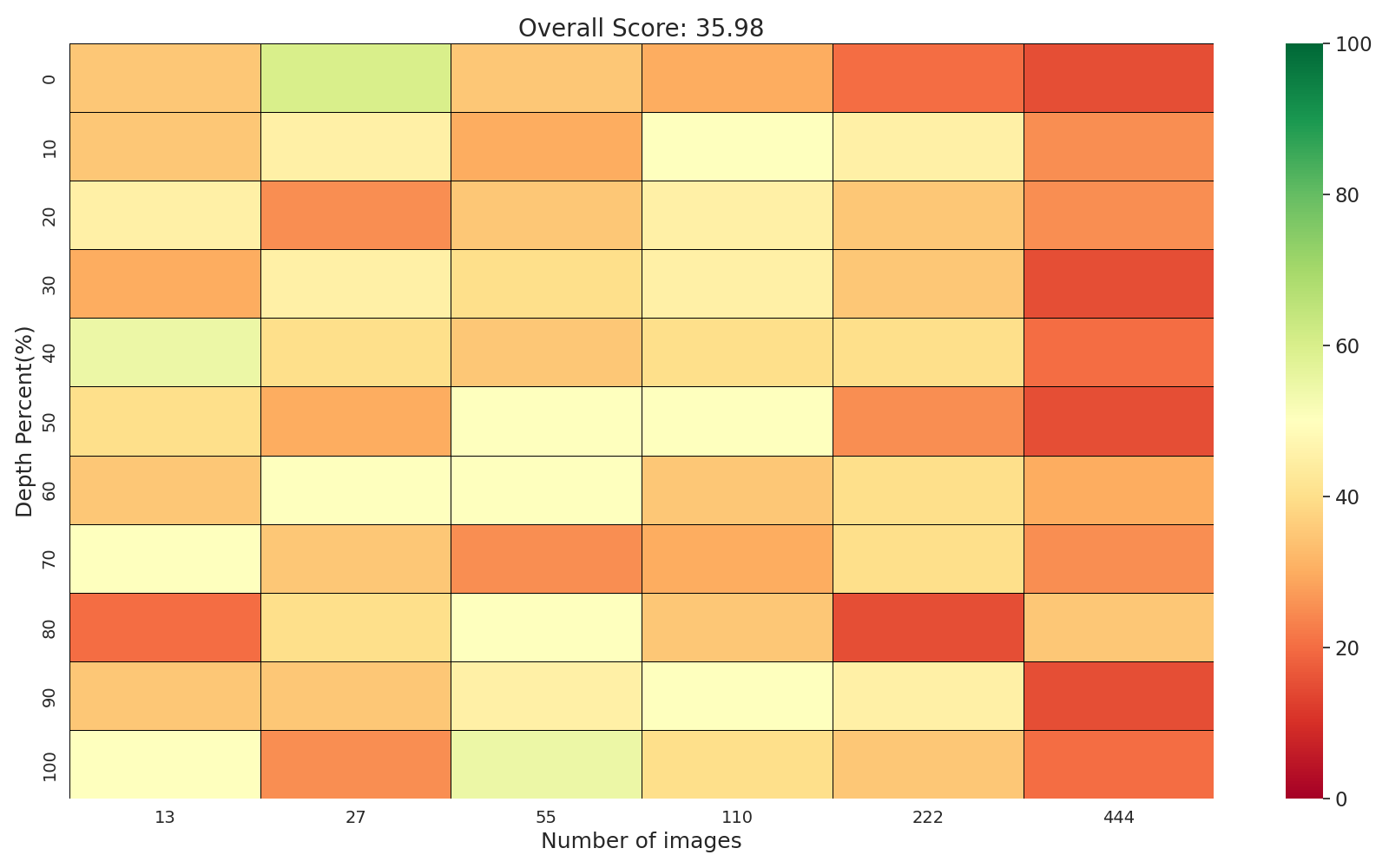}}
    \subfigure[GPT-4o]{\includegraphics[width=6.8cm]{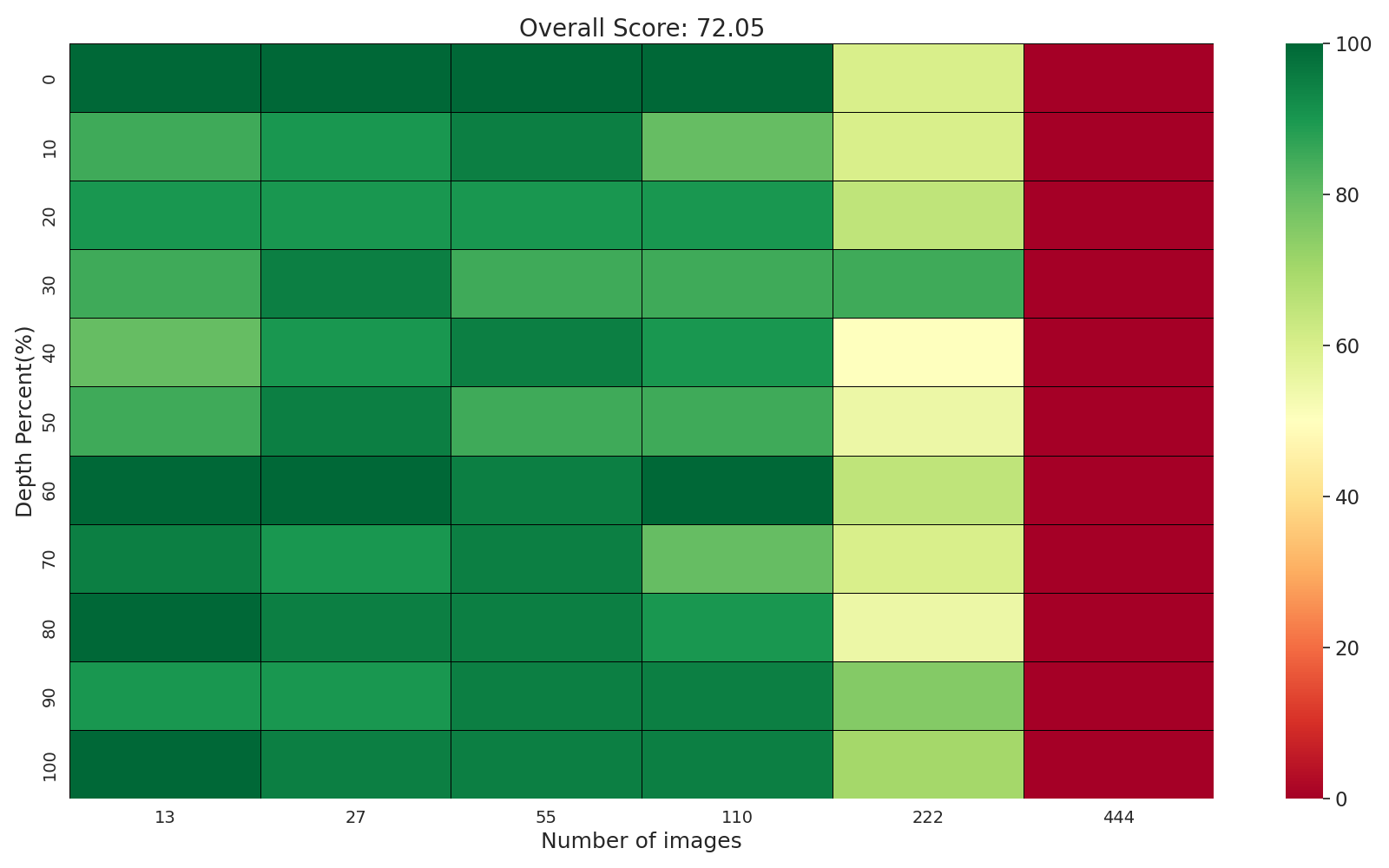}}
    \subfigure[Random]{\includegraphics[width=6.8cm]{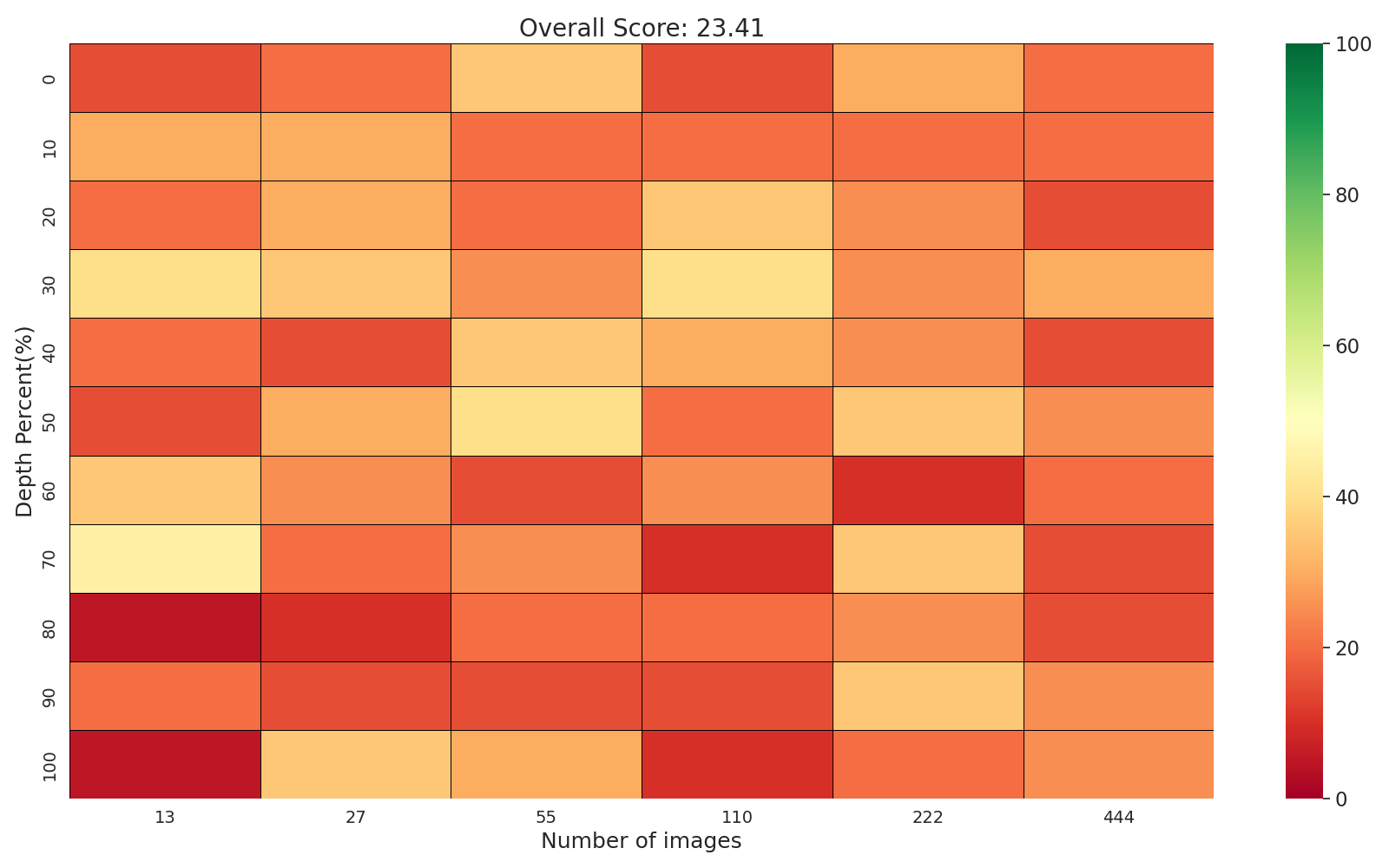}}
    \subfigure[OmChat]{\includegraphics[width=6.8cm]{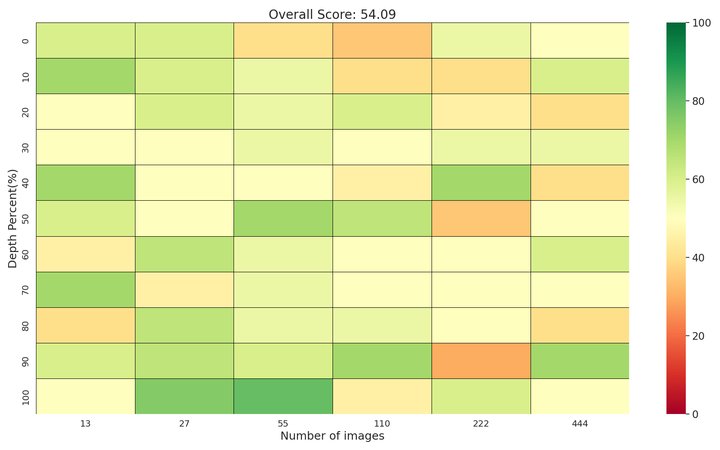}}
\caption{\textbf{TV Needle performance on LLaVa-1.5, GPT-4o, Random, and OmChat.}}
\label{tv_needle}
\end{figure}

We propose a benchmark dataset, named as Temporal Visual Needle in a Haystack (TV Needle), to assess the MLLMs' ability to comprehend temporal visual details within videos. Inspired by ``Needle in Haystack" \citep{LLMTest_NeedleInAHaystack}, we extend the concept of the needle task from text-based to a multimodal version. The TV Needle dataset retains the original objective of locating the ``text needle" information from long documents but introduces an additional challenge by requiring the identification of visual needles, including temporal information from long videos.

\begin{figure}[htbp]
  \centering
  \includegraphics[width=13cm]{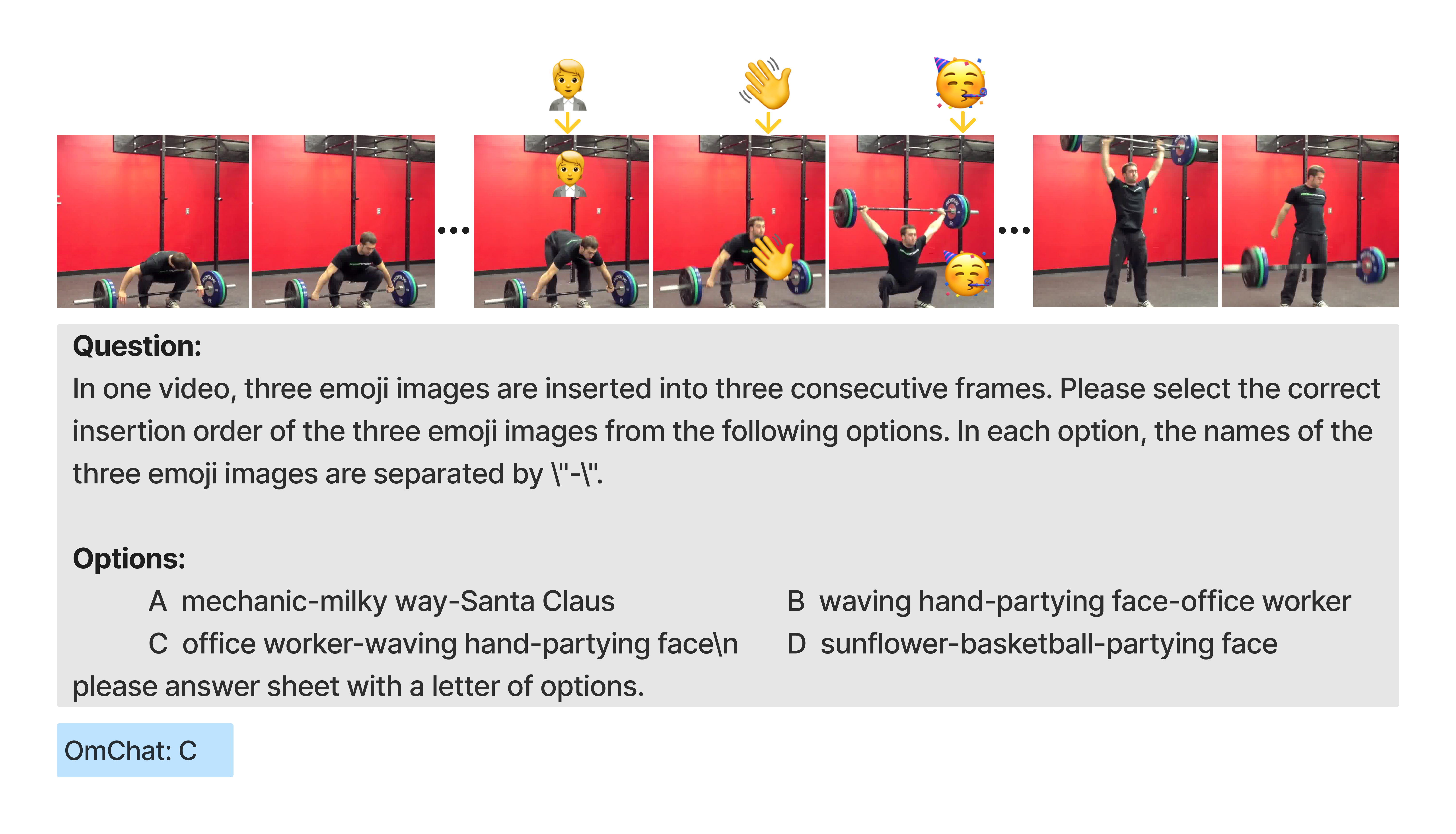}
  \caption{\textbf{Illustration of the TV Needle.} Three emojis are strategically embedded within three consecutive frames amidst a plethora of frames.}
  \label{video_needle}
\end{figure}

Specially, the long videos sourced from ActivityNet \citep{caba2015activitynet} are utilized as the ``haystack" for the TV Needle dataset. A subset of 10 to 16 videos is randomly selected and subsequently concatenated to create a single testing video. This process is repeated to yield a total of 20 distinct testing videos. This methodology ensures a comprehensive and varied evaluation of the system’s performance. These testing videos have duration ranging from 938 to 1,461 seconds. To create the ``needle" for the task, emojis are posted into the videos. Three emojis are randomly selected from a pool of 200 emojis and inserted into three consecutive frames within the video. This process introduces a temporal multimodal challenge where the model need to identify and locate these emoji sequences within the video content. Figure \ref{video_needle} provides an illustrative example of the TV Needle dataset, showcasing how the emojis are integrated into the video frames. Following the ``Needle in Haystack" approach \citep{LLMTest_NeedleInAHaystack}, emojis are inserted at varying depths within the video frames, ranging from 0\% to 100\%. 
% When quantified by the number of frames, this corresponds to a range from 13 frames to 444 frames in total.
We create subsets of varying number of frames, ranging from a minimum of 13 to a maximum of 444.
% This calculation is based on the context length of the model, which spans from 8k to 256k tokens, with each image contributing 576 tokens.
With each frame contributing 576 tokens, this corresponds an input length of 8k to 256k tokens for the MLLM being evaluated.
By incorporating emojis into the video frames at varying depths and challenging the model to locate and identify these sequences, the TV Needle dataset offers a unique and engaging task that measure the ability of MLLMs to understand and extract temporal visual details from videos.
% , enhancing its multimodal comprehension capabilities.

In Figure \ref{tv_needle}, we present a comparative analysis of the performance of OmChat, LLaVa-1.5, and GPT-4o. It is noteworthy that GPT-4o, despite being the top performer, does not perfectly solve the task. Its performance begins to significantly deteriorate on videos comprising 222 frames (equivalent to 128k tokens), and it fails to process 444 frames (256k tokens) due to API constraints.
Our model, OmChat, demonstrates consistent and commendable performance across all input lengths and needle positions. It surpasses both the random baseline and LLaVa-1.5 by a substantial margin. We attribute this robust capability of understanding long videos to our progressive training strategy, which gradually equipped our model with long context capability.

% {\color{red}{\textbf{Need table and explanation here}}}

\subsubsection{Multi-Image and Video Benchmarks}

% \begin{table}[ht]
% \centering
% \caption{Evaluation Results on Multi-image and Video Benchmarks}
% \resizebox{\textwidth}{!}{% Resize table to fit within the page width
% \begin{tabular}{ccccccc|c}
% \hline
% \textbf{Model} & \textbf{Model Type} & \textbf{Mantis-Eval} & \textbf{Q-Bench} & \textbf{MileBench Real.} & \textbf{MileBench Diag.} & \textbf{MVBench} & \textbf{Average}\\ \hline
% \begin{tabular}[c]{@{}c@{}} LLaVA-1.5-7B \end{tabular} & Single Image & 31.3 & 49.3 & 38.0 & 2.1 & 36.0 & 31.4 \\ \hline
% \begin{tabular}[c]{@{}c@{}} LLaVA-1.6-7B \end{tabular} & Single Image & 45.6 & 54.8 & 38.1 & 7.2 & 40.9 & 37.3 \\ \hline
% \begin{tabular}[c]{@{}c@{}} Qwen-VL-Chat \end{tabular} & Single Image & 39.2 & 45.9 & 39.1 & 37.2 & 42.2 & 40.7\\ \hline
% \begin{tabular}[c]{@{}c@{}} VILA \end{tabular} & Multi-image & 51.2 & 45.7 & 44.4 & 12.8 & 49.4 & 40.7\\ \hline
% \begin{tabular}[c]{@{}c@{}} Mantis-CLIP \end{tabular} & Multi-image & 55.8 & 66.0 & 47.5 & 19.9 & 48.3 & 47.5 \\ \hline
% \begin{tabular}[c]{@{}c@{}} OmChat-8B(Ours) \end{tabular} & Multi-image & \textbf{56.2} & \textbf{74.8} & \textbf{51.7} & \textbf{41.7} & \textbf{51.2} & \textbf{55.1} \\ \hline
% \begin{tabular}[c]{@{}c@{}} GPT-4V \end{tabular} & Multi-image & 62.7 & 76.5 & 53.0 & 99.4 & 43.5 & 67.0\\ \hline
% \end{tabular}
% }
% \label{tab:multi-image results}
% \end{table}

\begin{table}[ht]
\centering
\caption{\textbf{Evaluation results on multi-image and video benchmarks}}
\resizebox{\textwidth}{!}{% Resize table to fit within the page width
\begin{tabular}{ccccccc|c}
\toprule
\textbf{Model} & \textbf{Model Type} & \textbf{Mantis-Eval} & \textbf{Q-Bench} & \textbf{MileBench Real} & \textbf{MileBench Diag} & \textbf{MVBench} & \textbf{Avg}\\ 
\midrule
LLaVA-1.5-7B & Single Image & 31.3 & 49.3 & 38.0 & 2.1 & 36.0 & 31.4 \\ 
LLaVA-1.6-7B & Single Image & 45.6 & 54.8 & 38.1 & 7.2 & 40.9 & 37.3 \\ 
Qwen-VL-Chat & Single Image & 39.2 & 45.9 & 39.1 & 37.2 & 42.2 & 40.7\\  \midrule
VILA & Multi-image & 51.2 & 45.7 & 44.4 & 12.8 & 49.4 & 40.7\\ 
Mantis-CLIP & Multi-image & 55.8 & 66.0 & 47.5 & 19.9 & 48.3 & 47.5 \\ 
OmChat-8B(Ours) & Multi-image & \textbf{56.2} & \textbf{74.8} & \textbf{51.7} & \textbf{41.7} & \textbf{51.2} & \textbf{55.1} \\ 
GPT-4V & Multi-image & 62.7 & 76.5 & 53.0 & 99.4 & 43.5 & 67.0\\ 
\bottomrule
\end{tabular}
}
\label{tab:multi-image results}
\end{table}

In Table \ref{tab:multi-image results}, we compare our model, OmChat, on multi-image and video benchmarks with other state-of-the-art (SOTA) open-source and closed-source models. For MVBench, OmChat is evaluated using 16 frames to get the best performance.

%Mantis-Eval and Q-Bench are used to test model performance in multi-image scenarios, while MVBench evaluates video understanding ability using multiple frames as input. MileBench combines both multi-image and video tasks, challenging the model to understand the content of multiple images (semantic tasks) and to have strong video understanding capabilities (temporal tasks). {\color{red}{\textbf{I think we don't need data introduction here because we already introduced.}}}

Firstly, Comparing with models trained on single-image tasks, such as LLaVA-1.5 and Qwen-VL-Chat~\cite{qwen}, OmChat demonstrates a superior performance across all benchmarks, scoring more than 10 points higher on average. Additionally, when compared with SOTA multi-image models like Mantis~\cite{jiang2024mantis} and VILA~\cite{lin2024vila}, OmChat still outperforms them, likely due to our carefully designed multi-image and video data format.

Furthermore, when compared with the closed-source model GPT-4V, OmChat achieves a better score on MVBench, which requires a strong temporal video understanding capability.

\section{Ablation \& Analysis}
%\subsection{AnyRes Format Ablation}

\subsection{Multi-image and Video Format Ablation} \label{mutil-image ablation}

% \begin{table}[ht]
% \centering
% \caption{\textbf{Ablation Study on Multi-Image Data Format.} We compare five different data formats for handling multi-image and video data, labeled from F0 to F4. The symbols \texttt{<im\_start>} and \texttt{<im\_end>} are special tokens used as image delimiters. The term ``image/frame" indicates that the text prompt ``image" is used for multi-image data, while the text prompt ``frame" is used for video data.}
% \resizebox{\textwidth}{!}{% Resize table to fit within the page width
% \begin{tabular}{|c|c|c|c|c|c|c|c|}
% \hline
%  & \textbf{Format} & \textbf{Mantis-Eval} & \textbf{Q-Bench} & \textbf{MileBench Real.} & \textbf{MileBench Diag.} & \textbf{MVBench} & \textbf{Average}\\ \hline
% F0 & \begin{tabular}[c]{@{}c@{}}\texttt{<image><image>...}\end{tabular} & 41.0 & 47.3 & 45.8 & 32.5 & 42.0 & 41.7 \\ \hline
% F1 &\begin{tabular}[c]{@{}c@{}} \texttt{<im\_start><image><im\_end>} \end{tabular} & 39.2 & 50.8 & 45.9 & 34.7 & 44.7 & 43.1 \\ \hline
% F2 & \begin{tabular}[c]{@{}c@{}} \texttt{image \{i\}: <image> } \end{tabular} & 52.1 & 73.5 & 48.0 & 21.8 & 47.5 & 48.6\\ \hline
% F3 & \begin{tabular}[c]{@{}c@{}} \texttt{image \{i\}: <im\_start><image><im\_end>} \end{tabular} & 56.2 & 73.8 & 48.9 & 32.8 & 48.6 & 52.1 \\ \hline
% F4 & \begin{tabular}[c]{@{}c@{}} \texttt{image/frame \{i\}: <im\_start><image><im\_end>} \end{tabular} & \textbf{56.2} & \textbf{74.8} & \textbf{51.7} & \textbf{41.7} & \textbf{50.2} & \textbf{54.9} \\ \hline
% \end{tabular}
% }
% \label{tab:multi-image ablation}
% \end{table}

\begin{table}[ht]
\centering
\caption{\textbf{Ablation study on multi-image data format.} We compare five different data formats for handling multi-image and video data, labeled from F0 to F4. The symbols \texttt{<im\_start>} and \texttt{<im\_end>} are special tokens used as image delimiters. The term ``image/frame" indicates that the text prompt ``image" is used for multi-image data, while the text prompt ``frame" is used for video data.}
\renewcommand{\arraystretch}{1.5} % Set row height
\resizebox{\textwidth}{!}{%
\begin{tabular}{c|m{5cm}|>{\centering\arraybackslash}m{2cm} >{\centering\arraybackslash}m{2cm} >{\centering\arraybackslash}m{2cm} >{\centering\arraybackslash}m{2cm} >{\centering\arraybackslash}m{2cm} >{\centering\arraybackslash}m{2cm}}
\toprule
  & \textbf{Format} & \textbf{Mantis-Eval} & \textbf{Q-Bench} & \textbf{MileBench Real.} & \textbf{MileBench Diag.} & \textbf{MVBench} & \textbf{Average} \\ 
\midrule
F0 & \texttt{<image><image>...} & \large{41.0} & \large{47.3} & \large{45.8} & \large{32.5} & \large{42.0} & \large{41.7} \\ \hline
F1 & \texttt{<im\_start><image><im\_end>} & \large{39.2} & \large{50.8} & \large{45.9} & \large{34.7} & \large{44.7} & \large{43.1} \\ \hline
F2 & \texttt{image \{i\}: <image>} & \large{52.1} & \large{73.5} & \large{48.0} & \large{21.8} & \large{47.5} & \large{48.6}\\ \hline
F3 & \texttt{image \{i\}: <im\_start><image><im\_end>} & \large{56.2} & \large{73.8} & \large{48.9} & \large{32.8} & \large{48.6} & \large{52.1} \\ \hline
F4 & \texttt{image/frame \{i\}: <im\_start><image><im\_end>} & \large{\textbf{56.2}} & \large{\textbf{74.8}} & \large{\textbf{51.7}} & \large{\textbf{41.7}} & \large{\textbf{50.2}} & \large{\textbf{54.9}} \\ 
\bottomrule
\end{tabular}
}
\label{tab:multi-image ablation}
\end{table}

% \begin{table}[ht]
% \centering
% \caption{\textbf{Ablation Study on Multi-Image Data Format.} We compare five different data formats for handling multi-image and video data, labeled from F0 to F4. The symbols \texttt{<im\_start>} and \texttt{<im\_end>} are special tokens used as image delimiters. The term ``image/frame" indicates that the text prompt ``image" is used for multi-image data, while the text prompt ``frame" is used for video data.}
% \resizebox{\textwidth}{!}{% Resize table to fit within the page width
% \begin{tabular}{c|cccccc|c}
% \toprule
%  & \textbf{Format} & \textbf{Mantis-Eval} & \textbf{Q-Bench} & \textbf{MileBench Real.} & \textbf{MileBench Diag.} & \textbf{MVBench} & \textbf{Average} \\ 
% \midrule
% F0 & \begin{tabular}[c]{@{}c@{}}\texttt{<image><image>...}\end{tabular} & 41.0 & 47.3 & 45.8 & 32.5 & 42.0 & 41.7 \\ 
% F1 &\begin{tabular}[c]{@{}c@{}} \texttt{<im\_start><image><im\_end>} \end{tabular} & 39.2 & 50.8 & 45.9 & 34.7 & 44.7 & 43.1 \\ 
% F2 & \begin{tabular}[c]{@{}c@{}} \texttt{image \{i\}: <image> } \end{tabular} & 52.1 & 73.5 & 48.0 & 21.8 & 47.5 & 48.6\\ 
% F3 & \begin{tabular}[c]{@{}c@{}} \texttt{image \{i\}: <im\_start><image><im\_end>} \end{tabular} & 56.2 & 73.8 & 48.9 & 32.8 & 48.6 & 52.1 \\ 
% F4 & \begin{tabular}[c]{@{}c@{}} \texttt{image/frame \{i\}: <im\_start><image><im\_end>} \end{tabular} & \textbf{56.2} & \textbf{74.8} & \textbf{51.7} & \textbf{41.7} & \textbf{50.2} & \textbf{54.9} \\ 
% \bottomrule
% \end{tabular}
% }
% \label{tab:multi-image ablation}
% \end{table}

As shown in Table \ref{tab:multi-image ablation}, we conduct an ablation study to explore different input formats for improving the model's ability to handle multi-image and video scenarios. Various combinations of text prompts and special tokens are tested, with five different input formats labeled from F0 to F4. For this experiment, we use the same multi-image and single-image fine-tuning datasets and add an extra video instruction dataset for format F4.

Format F0 simply concatenates all the image tokens together, similar to how most single-image models operate. Despite this straightforward approach, our model using format F0 achieved higher scores on multi-image and video benchmarks than other single image models, benefiting from the utilization of interleaved data and multi-image and video data during the pretraining phase.

According to the experiment results, input formats F3 and F4, which use the text prompt ``image \{i\}" and special tokens \texttt{<im\_start>} and \texttt{<im\_end>} as image delimiters, proves to be effective for multi-image scenarios. Among these, format F4 delivers the best performance across all benchmarks.

A detailed analysis of the benchmark results for format F4 reveals that the text prompt ``image \{i\}" performs better for semantic multi-image tasks such as Mantis-Eval and Q-Bench. Additionally, the text prompt ``frame \{i\}" yields high scores on temporal video tasks, such as certain components in MileBench Realistic Evaluation and MVBench. This demonstrates that the combination of text prompts ``image \{i\}" and ``frame \{i\}" is an effective approach to help the model understand both multi-image and video scenarios.

\subsection{Data Selection Ablation}
\label{data selection ablation}
% \begin{table}[ht]
% \centering
% \caption{\textbf{Evaluation Results on Continuous Training for Fine-tuning Data Selection.} The evaluation results for models trained with varying data portions using a continuous training strategy are presented. The data portion size is determined based on the total number of tokens in the answers from the fine-tuning dataset. The total average scores encompass performance metrics from both text benchmarks and multimodal benchmarks.}
% \begin{tabular}{|c|c|c|}
% \hline
% \textbf{Data Portion Size}  & \textbf{Total Average} \\ \hline
% \begin{tabular}[c]{@{}c@{}}\texttt{10M}\end{tabular} & 56.05 \\ \hline
% \begin{tabular}[c]{@{}c@{}} \texttt{20M} \end{tabular}  & \textbf{57.09}  \\ \hline
% \begin{tabular}[c]{@{}c@{}} \texttt{40M} \end{tabular}  & 56.59 \\ \hline
% \begin{tabular}[c]{@{}c@{}} \texttt{60M} \end{tabular}  & 56.64 \\ \hline
% \begin{tabular}[c]{@{}c@{}} \texttt{80M} \end{tabular} & 56.01 \\ \hline
% \end{tabular}
% \label{tab:sft data selection}
% \end{table}

\begin{table}[ht]
\centering
\caption{\textbf{Evaluation results on continuous training for instruction tuning Ddata selection.} The evaluation results for models trained with varying data portions using a continuous training strategy are presented. The data portion size is determined based on the total number of tokens in the answers from the fine-tuning dataset. The total average scores encompass performance metrics from both text benchmarks and multimodal benchmarks.}
\begin{tabular}{cc}
    \toprule
    \textbf{Data Portion Size} & \textbf{Total Average} \\
    \midrule
    \texttt{10M} & 56.05 \\
    \texttt{20M} & \textbf{57.09} \\
    \texttt{40M} & 56.59 \\
    \texttt{60M} & 56.64 \\
    \texttt{80M} & 56.01 \\
    \bottomrule
    \end{tabular}
    \label{tab:sft data selection}
\end{table}

To validate the effectiveness of our continuous training strategy and select an optimal combination of fine-tuning datasets, we conduct fine-tuning experiments using the approach introduced in Section 2.5. Initially, a large dataset comprising 0.2B tokens in answers is sampled based on manually designed percentages. This large dataset is then divided into multiple data portions, each maintaining the designed percentages.

We use our language model, OmBase, as the base for MLLM and perform a data selection ablation. After applying the continuous training strategy to these data portions, we evaluate the fine-tuned models on multiple multimodal and text-only benchmarks, using a total average score to compare performances.

%
% For multimodal benchmarks, we selected ScienceQA, TextVQA, 
% MMBench, OKVQA, MMMU, AI2D, and ChartQA. For text-only benchmarks, % we used Ceval, MMLU, and GSM8K. 
%

As shown in Table \ref{tab:sft data selection}, the total average score increases from the data portion of 10M to 20M tokens, indicating that adding more samples from each dataset improves model performance. However, from the data portion of 40M to 80M tokens, the total average score decreases. Detailed analysis of each benchmark's scores reveals that the decline is primarily in text-only benchmarks. This drop in text performance may be due to excessive training steps on multimodal tasks. Conversely, the multimodal scores only slightly increase with data portions from 40M to 80M tokens.Therefore, considering the trade-off between multimodal and text-only capabilities, we select the data portion size of 20M tokens as the best instruction tuning data combination.

\subsection{Qualitative Examples}
% \begin{figure}[htbp]
%   \centering
%   \includegraphics[width=14cm]{error_analysis_v3.png}
%   \caption{\textbf{Qualitative comparison samples of OmChat against open source models.}}
%   \label{errors}
% \end{figure}

\begin{figure}[ht]
  \centering
  \includegraphics[width=14cm]{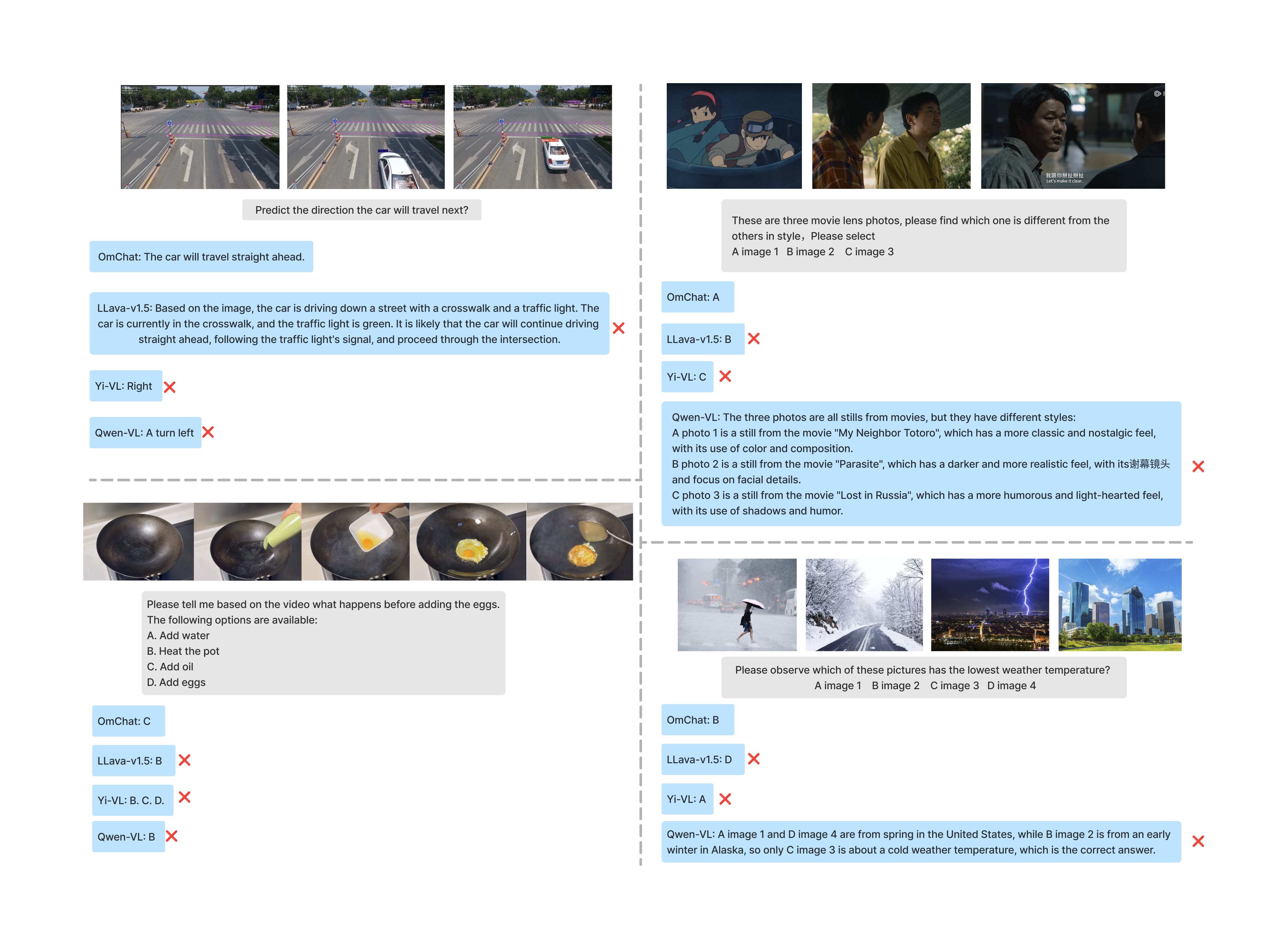}
  \caption{\textbf{Qualitative comparison samples of OmChat against open source models.}}
  \label{errors}
\end{figure}

In our study, we present a series of examples derived from benchmarks, which include multi-image or video data (displayed via extracted frames) as shown in Figure \ref{errors}. The OmChat model demonstrates superior performance in comparison to other robust MLLMs. Its proficiency is particularly evident in its ability to predict future actions in a video, make judgements rooted in common sense, and compare visual features across consecutive images. The emergent capability of the OmChat model to understand long-context visuals broadens the scope of its potential applications significantly. This underscores the model’s versatility and effectiveness, positioning it as a leading tool in the field of multimodal large language models. More examples of OmChat for various visual tasks are depicted in Figure \ref{show1} and Figure \ref{show2} in Appendix. 

% \begin{figure}[htbp]
%   \centering  \includegraphics[width=14.5cm, height=21cm]{show_case_v11_2.png}
%   \caption{Examples of OmChat Handling Various Tasks}
%   \label{show1}
% \end{figure}

% \begin{figure}[htbp]
%   \centering  \includegraphics[width=14.5cm, height=21cm]{show_case_v11_1.png}
%   \caption{More Examples of OmChat Handling Various Tasks}
%   \label{show2}
% \end{figure}

\section{Conclusion}
This paper introduced OmChat, a multimodal model designed for handling long contexts and video understanding tasks. OmChat’s active progressive multimodal pretraining strategy, combined with high-quality supervised fine-tuning datasets and support for  any-aspect high image resolutions, ensures exceptional performance across various benchmarks.

OmChat performs outstandingly in tasks involving multiple images and videos, managing complex visual data effectively and supporting a context length of up to 512K tokens. Our analysis highlights the importance of higher image resolutions, progressive multimodal pretraining, and high-quality training data in achieving state-of-the-art performance.

OmChat sets a new benchmark for multimodal large language models. Future work will enhance its capabilities, explore efficient training techniques, and expand its application to more multimodal tasks.

\bibliography{iclr2023_conference}
\bibliographystyle{iclr2023_conference}

\appendix
\section{Appendix}

\begin{figure}[htbp]
  \centering  \includegraphics[width=14.5cm, height=21cm]{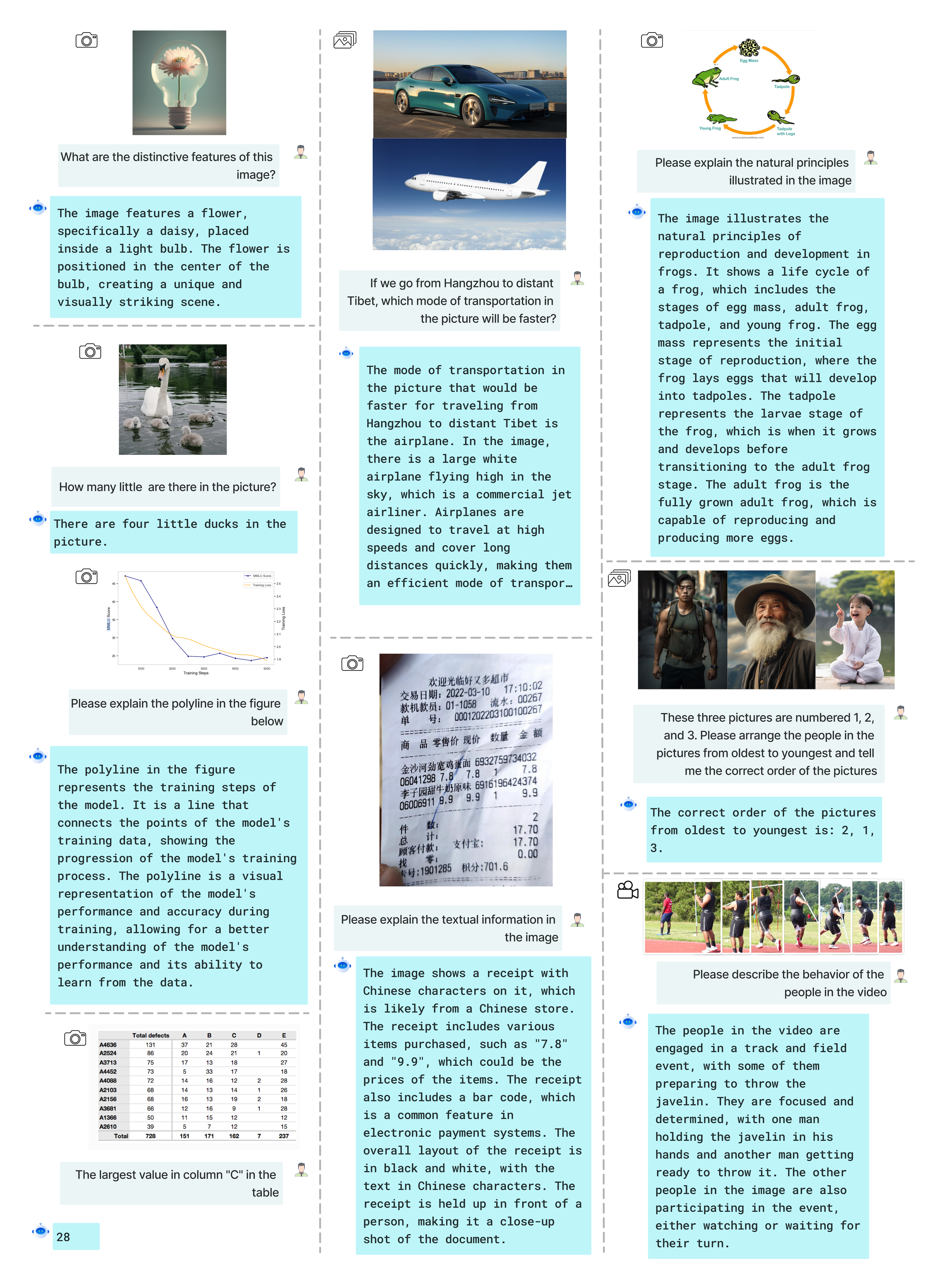}
  \caption{\textbf{Examples of OmChat Handling Various Tasks}}
  \label{show1}
\end{figure}

\begin{figure}[htbp]
  \centering  \includegraphics[width=14.5cm, height=21cm]{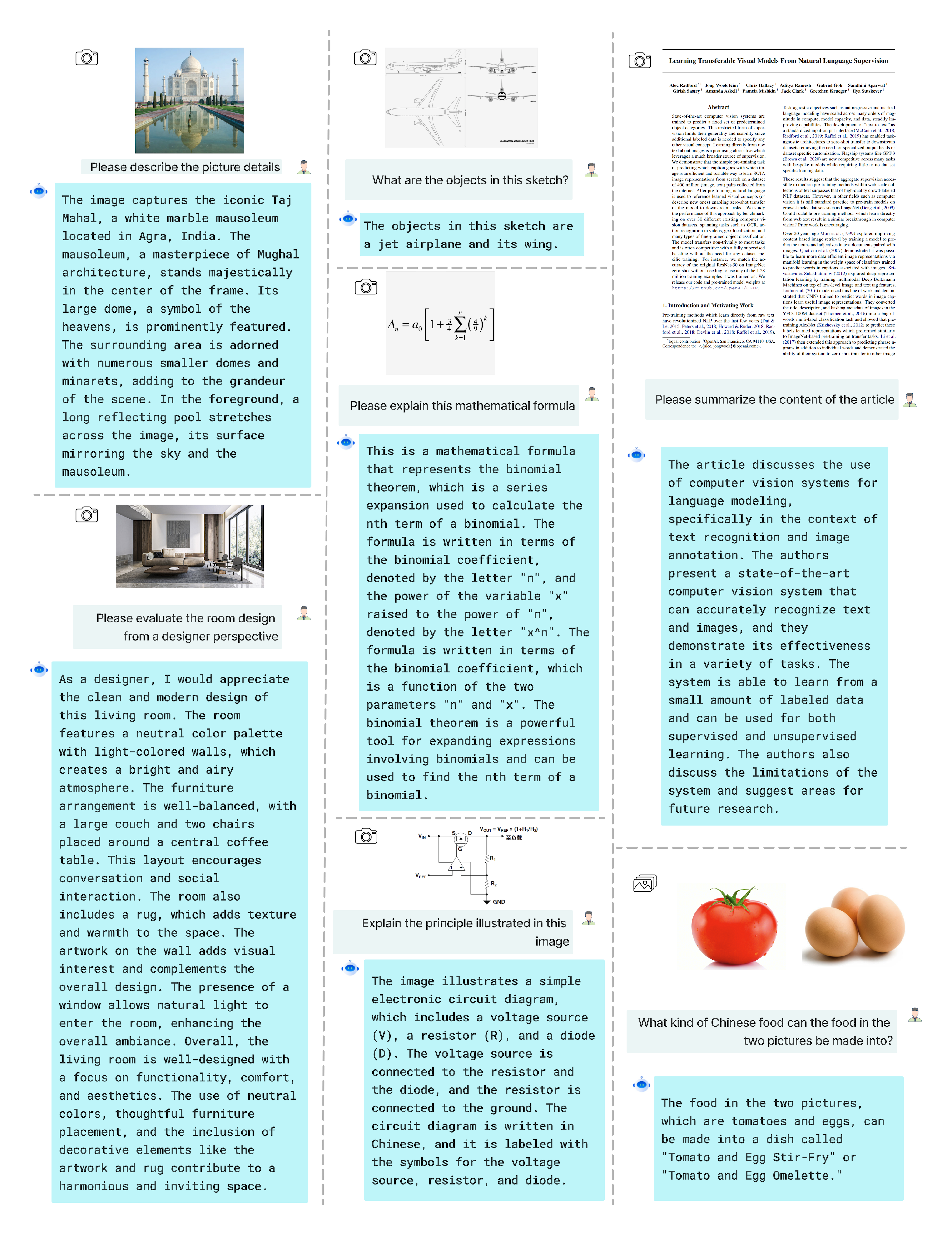}
  \caption{\textbf{More Examples of OmChat Handling Various Tasks}}
  \label{show2}
\end{figure}
% You may include other additional sections here.

\end{document}